\documentclass[10pt,twocolumn,letterpaper]{article}
\usepackage{authblk}

\usepackage[pagenumbers]{cvpr} 

\usepackage[dvipsnames]{xcolor}
\usepackage{ulem}

\usepackage{tikz}
\usepackage{pgfplots}

\usepackage{amsmath, booktabs, caption, longtable}

\definecolor{cvprblue}{rgb}{0.21,0.49,0.74}

\usepackage[pagebackref,breaklinks,colorlinks,citecolor=cvprblue]{hyperref}

\usepackage{float}

\pgfplotsset{compat=1.18}

\usepackage[nolist]{acronym}

\begin{acronym}
    \acro{AP}{Average Precision}
    \acro{ADC}{Apparent diffusion coefficient}
    \acro{ASSD}{Average Symmetric Surface Distance}
    \acro{AUC}{Area Under the Curve}
    \acro{BraTS-M}{ASNR-MICCAI BraTS Brain Metastasis Challenge}
    \acro{BraTS-G}{RSNA-ASNR-MICCAI BraTS Continuous Evaluation Challenge}
    \acro{CNN}{Convolutional Neural Network}
    \acro{CCA}{Connected Component Analysis}
    \acro{CT}{Computed Tomography}
    \acro{clDice}{Centerline Dice Similarity Coefficient}
    \acro{DL}{Deep Learning}
    \acro{DQ}{Detection Quality}
    \acro{TP}{True Positive}
    \acro{FP}{False Positive}
    \acro{FN}{False Negatives}
    \acro{DSC}{Dice Similarity Coefficient}
    \acro{DWI}{Diffusion Weighted Imaging}
    \acro{ET}{Enhancing Tumor}
    \acro{GAN}{Generative Adversarial Network}
    \acro{gvDSC}{global volumetric Dice Similarity Coefficient}
    \acro{HD}{Hausdorff Distance}
    \acro{ISLES}{Ischemic Stroke Lesion Segmentation 2022 challenge}
    \acro{IOU}{Intersection over Union}
    \acro{InS}{Instance Segmentation}
    \acro{ML}{Machine Learning}
    \acro{MM}{Matching Metric}
    \acro{MONAI}{Medical Open Network for Artificial Intelligence}
    \acro{MRI}{Magnetic Resonance Imaging}
    \acro{MS}{Matching Score}
    \acro{NETC}{Nonenhancing Tumor Core}
    \acro{NN}{Neural Network}
    \acro{NSD}{Normalized Surface Dice}
    \acro{PGT}{Peak Ground Truth}
    \acro{PQ}{Panoptic Quality}
    \acro{RQ}{Recognition Quality}
    \acro{RWMP}{Real World Model Performance}
    \acro{SQ}{Segmentation Quality}
    \acro{PanS}{Panoptic Segmentation}
    \acro{SNFH}{Surrounding Non-enhancing FLAIR Hyperintensity}
    \acro{SemS}{Semantic Segmentation}
    \acro{SQASSD}{Segmentation Quality Average Symmetric Surface Distance}
    \acro{t1c}{post-contrast T1-weighted}
    \acro{t1w}{pre-contrast T1-weighted}
    \acro{t2f}{T2-weighted Fluid Attenuated Inversion Recovery}
    \acro{FLAIR}{T2-weighted Fluid Attenuated Inversion Recovery}
    \acro{t2w}{T2-weighted}
    \acro{VerSe}{Large Scale Vertebrae Segmentation Challenge}

\end{acronym}

\newcommand{\eac}[1]{\textit{\ac{#1}}}

\newcommand{\eacf}[1]{\textit{\acf{#1}}}

\newcommand{\eacs}[1]{\textit{\acs{#1}}}

\newcommand{\eacl}[1]{\textit{\acl{#1}}}

\usepackage{tabularx}

\title{\textit{panoptica} -- instance-wise evaluation of \\3D semantic and instance segmentation maps}

\author[1,2,3,4]{Florian Kofler} 
\author[2, 4]{Hendrik Möller} 
\author[5]{Josef A. Buchner}
\author[6,2]{Ezequiel de la Rosa}
\author[2,3]{Ivan Ezhov} 
\author[2,1]{\\Marcel Rosier}
\author[1]{Isra Mekki}
\author[2]{Suprosanna Shit}
\author[7]{Moritz Negwer}
\author[2,7]{Rami Al-Maskari}
\author[7,8,9,10]{\\Ali Ertürk}
\author[11,12,13]{Shankeeth Vinayahalingam}
\author[14,15]{Fabian Isensee} 
\author[16,2]{Sarthak Pati}
\author[2,17]{\\Daniel Rueckert}
\author[4]{Jan S. Kirschke} 
\author[18]{Stefan K. Ehrlich} 
\author[19]{Annika Reinke}
\author[20]{Bjoern Menze} 
\author[4]{\\Benedikt Wiestler}
\author[1]{Marie Piraud}

\affil[1]{\footnotesize Helmholtz AI, Helmholtz Munich, Neuherberg, Germany}
\affil[2]{\footnotesize Department of Computer Science, TUM School of Computation, Information and Technology, Technical University of Munich, Munich, Germany}
\affil[3]{\footnotesize TranslaTUM - Central Institute for Translational Cancer Research, Technical University of Munich, Munich, Germany}
\affil[4]{\footnotesize Department of Diagnostic and Interventional Neuroradiology, School of Medicine, Klinikum rechts der Isar, Technical University of Munich, Munich, Germany}
\affil[5]{\footnotesize Department of Radiation Oncology, School of Medicine, Klinikum rechts der Isar, Technical University of Munich, Munich, Germany}
\affil[6]{\footnotesize icometrix, Leuven, Belgium}
\affil[7]{\footnotesize Institute for Tissue Engineering \& Regenerative Medicine (iTERM), Helmholtz Munich, Neuherberg, Germany}
\affil[8]{\footnotesize Institute for Stroke and dementia research (ISD), University Hospital, LMU Munich, Munich, Germany}
\affil[9]{\footnotesize Graduate school of neuroscience (GSN), Munich, Germany}
\affil[10]{\footnotesize Munich cluster for systems neurology (Synergy), Munich, Germany}
\affil[11]{\footnotesize Department of Oral and Maxillofacial Surgery, Radboud University Medical Center Nijmegen, Nijmegen, The Netherlands}
\affil[12]{\footnotesize Department of Artificial Intelligence, Radboud University Medical Center Nijmegen, Nijmegen, The Netherlands}
\affil[13]{\footnotesize Department of Oral and Maxillofacial Surgery, University Hospital Münster, Münster, Germany}
\affil[14]{\footnotesize Applied Computer Vision Lab, Helmholtz Imaging, Hamburg, Germany}
\affil[15]{\footnotesize Division of Medical Image Computing, German Cancer Research Center (DKFZ), Heidelberg, Germany}
\affil[16]{\footnotesize Department of Pathology and Laboratory Medicine, Indiana University School of Medicine, Indianapolis, USA}
\affil[17]{\footnotesize Imperial College London, London, United Kingdom}
\affil[18]{\footnotesize Healthcare Technologies, SETLabs Research GmbH, Munich, Germany}
\affil[19]{\footnotesize Intelligent Medical Systems (IMSY) and HI Helmholtz Imaging, German Cancer Research Center (DKFZ), Heidelberg University, Heidelberg, Germany}
\affil[20]{\footnotesize Department of Quantitative Biomedicine, University of Zurich, Zurich, Switzerland}

\begin{document}

\maketitle
\vspace{-1cm}

\begin{abstract}
    This paper introduces \textit{panoptica}, a versatile and performance-optimized package designed for computing instance-wise segmentation quality metrics from 2D and 3D segmentation maps.
    \textit{panoptica} addresses the limitations of existing metrics and provides a modular framework that complements the original \eacs{IOU}-based \eacl{PQ} with other metrics, such as the distance metric \eacs{ASSD}.
    The package is open-source, implemented in Python, and accompanied by comprehensive documentation and tutorials.
    \textit{panoptica} employs a three-step metrics computation process to cover diverse use cases.
    We demonstrate the efficacy of \textit{panoptica} on various real-world biomedical datasets, where an instance-wise evaluation is instrumental for an accurate representation of the underlying clinical task.
    Overall, we envision \textit{panoptica} as a valuable tool facilitating in-depth evaluation of segmentation methods.
\end{abstract}

\section{Introduction}
\label{sec:introduction}
Segmentation is mostly formulated as a task related to either \textit{stuff}, representing all pixels/voxels belonging to a single class with the same label (\eac{SemS}) or \textit{things}, labeling each pixel/voxel of a class (e.g., metastatic tumor lesions) with an individual label (\eac{InS}) \citep{kirillov2019panoptic}.

For many biomedical segmentation problems, an instance-wise evaluation focusing on individual lesions is highly relevant and desirable for both scientific research and clinical applications.
This significance is evident in the context of multiple sclerosis, an inflammatory disease characterized by a variable number of white matter lesions in the brain.
In the prediction of disease activity, a graph neural network leveraging representations from individual lesions has been suggested \citep{prabhakar2023self}.
From a clinical perspective, the occurrence of a single new inflammatory lesion is crucial in establishing dissemination in time, playing a pivotal role in the diagnosis and ongoing monitoring of multiple sclerosis patients \citep{thompson2018diagnosis}.
Both the scientific modeling of disease activity as well as the clinical monitoring of patients are clearly dependent on accurate lesion-wise \eac{InS}.

Despite the apparent need for reliable instance-wise segmentation, many biomedical segmentation problems are still addressed as \eac{SemS} problems, partly due to the lack of appropriate instance labels and partly due to the advances in semantic segmentation methodology:


The biomedical image analysis community has successfully developed a plethora of (mostly \textit{U-Net}-based \citep{ronnebergerUnet,isensee2021nnu}) methods for solving the problem of \eac{SemS} across a range of imaging modalities in diverse applications such as primary brain tumors \citep{bakas2019identifying, kofler2021we,kofler2020brats}, brain metastases \citep{buchner2023development, buchner2023identifying}, vascular lesions \citep{sudre2022valdo},  multiple sclerosis \citep{kofler2023blob} or stroke \citep{hernandez2022isles} in brain MRI, liver tumors \citep{bilic2023liver, kofler2023blob} or the spine \citep{sekuboyina2021verse} in CT or light-sheet microscopy tasks \citep{kaltenecker2023virtual,bhatia2022spatial}.


In view of the need for reliable instance-wise segmentation outlined above, there have been attempts to optimize instance awareness in \eac{SemS}, e.g., \citep{kofler2023blob}.
Reuniting \textit{stuff} and \textit{things} by simultaneously evaluating \eac{SemS} and \eac{InS} is, therefore, a relevant but unmet need, especially in biomedical image analysis.



\noindent\textbf{Contribution:}
We introduce \textit{panoptica}, a modular, performance-optimized package to compute instance-wise segmentation quality metrics for 2D and 3D semantic- and instance segmentation maps.
On an instance level, \textit{panoptica} allows reporting detection metrics, such as  \eac{RQ} (equivalent to \textit{F1-score}), and segmentation quality metrics, such as \eac{SQ} and \eac{PQ}.
Furthermore, additional metrics such as \eac{ASSD} are available.
\textit{panoptica} is an open-source Python package, and comprehensive documentation and tutorials are available (\cf \Cref{sec:methods}).

To overcome limitations in the \ac{PQ} metric, which may obscure information related to either detection or segmentation quality \cite{reinke2021common, maierhein2023metrics},
\textit{panoptica} focuses on presenting all aspects relevant to instance segmentation tasks by complementing \eac{PQ} with other pertinent performance metrics.

The metric computation process in \textit{panoptica} is divided into three steps, offering flexibility for various use cases, \cf \Cref{sec:methods,sec:experiments}.
In contrast to existing solutions, \textit{panoptica} facilitates the evaluation of semantic segmentation methods at an instance level.
We showcase the capabilities of \textit{panoptica} across multiple biomedical data sets featuring real-world segmentation challenges.
For these experiments, instance-wise evaluation proves crucial in capturing the nuances of the underlying biomedical tasks. 
The instance-wise analysis reveals aspects of segmentation quality that may remain concealed when solely evaluating the entire image domain,\cf \Cref{sec:experiments}.

\section{Related Work}
\citet{kirillov2019panoptic} introduce the task of \eacf{PanS} as combination of \eacf{SemS} and \eacf{InS}.
They define a \eac{PanS} task as both assigning each pixel to a \textit{class-ID} and an \textit{instance ID}, for \eac{SemS} and \eac{InS} respectively.
Further, they propose the \eacf{PQ} metric, \cf \Cref{sec:evaluator}.
Although the metric was originally proposed for \eac{PanS} tasks, the \textit{Metrics Reloaded initiative} \cite{maierhein2023metrics} recommended employing the metric also for \eac{InS} tasks.
In this context, PQ allows capturing both detection and segmentation quality in one score and is an alternative to the $F\beta$-score assessing per-class detection quality. \\

The \eac{PQ} metric is implemented as part of the \href{https://github.com/Project-MONAI/MONAI/blob/dev/monai/metrics/panoptic_quality.py}{\eac{MONAI} framework} \citep{cardoso2022monai} and \href{https://torchmetrics.readthedocs.io/en/stable/detection/panoptic_quality.html}{torchmetrics} \citep{detlefsen2022torchmetrics}.
However, these implementations only support 2D and not 3D input data, and the \textit{torchmetrics} implementation provides no means for instance matching.
Therefore, it is only applicable to already matched instance maps.
Further, the \textit{Metrics Reloaded initiative} \cite{maierhein2023metrics} provides a 3D implementation in a \href{https://github.com/Project-MONAI/MetricsReloaded/blob/main/MetricsReloaded/processes/mixed_measures_processes.py}{side package} of \eac{MONAI}.
While the program can compute \eac{PQ}, \eac{SQ}, and \eac{RQ}, it is designed to always return a single metric.
Hence, multiple calls are necessary if all three metrics are required.
This is a conscious design choice as the program assumes that the user selects the most suitable metrics via the \href{https://metrics-reloaded.dkfz.de/}{Metrics Reloaded framework} \citep{maierhein2023metrics}.
Naturally, the computational overhead of multiple calls leads to a massive accumulation of processing times, prohibiting the evaluation of large-scale data sets.
Further, even though computed internally, the package offers no means to return \eac{TP}, \eac{FP}, \eac{FN}.
Unlike \textit{panoptica}, the implementation of the \textit{Metrics Reloaded initiative}, lacking instance approximation capabilities, cannot calculate instance-wise metrics for semantic inputs.

\citet{jungo2021pymia} implement a multitude of segmentation quality metrics in \textit{pymia} \citep{jungo2021pymia}; however, the package operates only on whole images and offers no means, for instance, approximation or matching.

\citet{chen2023sortedap} provide a good overview of \eac{InS} metrics and introduce a new metric called \textit{SortedAP}, which strictly decreases with both
object- and pixel-level imperfections.

\clearpage
\section{Methods}
\label{sec:methods}
\begin{figure*}[htbp]
  \centering
  \includegraphics[width=1.0\textwidth]{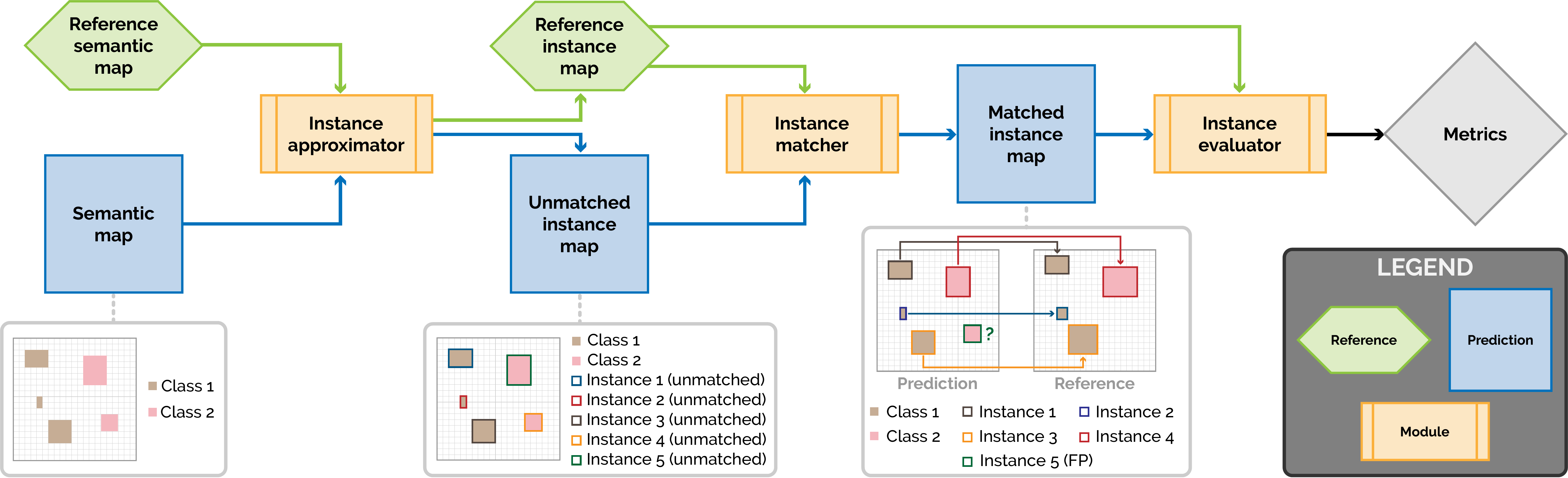}
  \caption{    \textit{panoptica}'s modular architecture.
    Factual \textit{ground truth} is often absent in human-annotated segmentation tasks \citep{kofler2023approaching}.
    Therefore, \textit{panoptica} operates on the concept of comparing \textit{model predictions} to \textit{reference labels}.
    The first module, \textit{Instance Approximator}, approximates instances; it computes instance maps from both reference and prediction semantic segmentation maps.
    The second module, \textit{Instance Matcher}, matches an \textit{unmatched instance map} to a \textit{reference instance map}; it outputs a \textit{matched instance map}.
    The third module, \textit{Evaluator}, computes instance-wise metrics by comparing the \textit{reference instance map} to a \textit{matched instance map}.
    \textit{panoptica}'s modular architecture allows the assembly of metric evaluation pipelines covering various use cases, \cf \Cref{sec:methods,sec:experiments}.
  }
  \label{fig:workflow}
\end{figure*}

\textit{panoptica} supports computing instance-wise segmentation, and object detection metrics from \textit{semantic}-, as well as \textit{unmatched} and \textit{matched} segmentation maps, \cf \Cref{fig:workflow}.

\textit{panoptica} separates the metric computation process into three steps.
This allows building modular metric computation pipelines, as illustrated in \Cref{fig:workflow}.
Due to its modular software architecture, resembling this three-step process, \textit{panoptica} can easily be extended with additional algorithms for instance approximation, - matching, and metrics.
To this end, a framework that automatically validates each input and output and defines all datatypes to be used is established.
Therefore, a user extending \textit{panoptica} with a custom algorithm receives immediate and understandable feedback regarding framework fit.

\textit{panoptica} is an open-source package written in Python and included in the
\href{https://github.com/BrainLesion}{BrainLes}
project.
It can be directly installed from
\href{https://pypi.org/project/panoptica/}{PyPI}
, and the source code is available on
\href{https://github.com/BrainLesion/panoptica/tree/main/panoptica}{GitHub}
.
Further, \textit{panoptica} is designed to process large-scale data-sets.
To achieve fast computation times, it implements parallel computations by leveraging \textit{Python}'s native \href{https://docs.python.org/3/library/multiprocessing.html}{multiprocessing} library.
Additionally, as many segmentation tasks only contain small foreground areas in an otherwise large image, inputs are cropped down to the combined bounding box of both prediction and reference to speed up computations.
Detailed performance reports are available in the Supplemental Material, \cf \Cref{sec:times}.

The
\href{https://github.com/BrainLesion}{BrainLes}
project features several
\href{https://github.com/BrainLesion/tutorials/tree/main/panoptica}{Jupyter notebook tutorials} to illustrate the different use cases for \textit{panoptica} with instance- and semantic segmentation maps.
Further, Supplemental Material \Cref{sec:curves} illustrates how similarity metric curves can be analyzed in dependency of \eac{IOU} thresholds.

\subsection{Instance approximation}
\label{sec:approximation}
While instance-wise metrics can directly be calculated from instance segmentation maps, semantic segmentation maps, by nature, do not contain instance information.
Therefore, at the time of release, the \textit{instance approximator} module provides means to approximate instances via \eac{CCA}.
For this \eac{CCA} users can select between \href{https://pypi.org/project/connected-components-3d/}{cc3d} \citep{silversmith_2021_5719536} and \href{https://docs.scipy.org/doc/scipy/reference/generated/scipy.ndimage.label.html}{scipy} backends.
For most hardware configurations, \textit{cc3d} is expected to perform better on 3D images, while \textit{scipy} promises better 2D performance.
Therefore, without user input, \textit{panoptica} defaults to the respective backend for 2D and 3D inputs.
For a detailed benchmark of \eac{CCA} backends, \cf Supplemental Material \Cref{sec:approximation}.

\subsection{Instance matching}
\label{sec:matching}
The \textit{instance matcher} module offers means to match instances based on a specified overlap metric.
At the time of this writing, we support \eac{IOU} and the \eac{DSC}.
For simplicity of explanation, we will denote the selected metric as \eac{MM}, its value as \eac{MS}.
The task of the \textit{instance matcher} is to assign each prediction instance to either a reference instance or as \eac{FP}.
Each \eac{FP} is relabeled to an unused label so it persists in the map.

We only ever match instances of the prediction to the reference, leaving any given reference map untouched.
Depending on the algorithm, we do support many-to-one matching, meaning multiple prediction instances can be matched to the same reference instance.

At the time of release, we offer two distinct instance-matching algorithms.
However, the software architecture allows to easily extend \textit{panoptica} with more sophisticated matching algorithms.

\subsubsection{Naive Threshold Matcher}
The \textit{naive threshold matcher} assigns all prediction instances that exceed a certain user-given \eac{MS} threshold with the reference instance.

If multiple predictions exceed the threshold for the same reference instance, only the one with the highest \eac{MS} is matched to that reference.
If the user specified that multiple predictions are allowed to be matched to the same reference instance, each of those matches are kept instead.
Should a prediction instance exceed the threshold for multiple reference instances, it is only mapped to the reference instance with the highest \eac{MS}.

\subsubsection{Maximize Many-to-One Matcher}
This \textit{many-to-one matching} algorithm tries to maximize the \eac{MM} by allowing multiple prediction instances to be assigned to the same reference instance if and only if the combined prediction instances have a greater \eac{MS} than the individual instances.
To this end, for each reference instance, it finds the best prediction instance based on the \eac{MM} and then tests for all other instances that have an \eac{MS} greater zero if including it would increase the score.
Therefore, this algorithm maximizes the individual specified \eac{MM} when matching but does not provide an overall optimal \eac{MS} solution across all instances.

\newpage
\subsection{Instance evaluation}
\label{sec:evaluator}
The \textit{instance evaluator} module calculates instance-wise metrics by comparing \textit{reference} to \textit{matched instance} segmentation maps, \cf \Cref{fig:workflow}.

\citet{kirillov2019panoptic} originally define the \acf{PQ} metric as \Cref{equation:pq_combined,equation:pq_separate}:

\begin{equation}
    \begin{aligned}
        \text{\textit{PQ}} & = \frac{\sum_{(R, P)\in \text{TP}} \text{\textit{f}}(R, P)}{\vert \text{TP} \vert + 0.5 \cdot \vert \text{FP} \vert + 0.5 \cdot \vert \text{FN} \vert}
    \end{aligned}
    \label{equation:pq_combined}
\end{equation}

\begin{equation}
    \begin{aligned}
        \text{\textit{PQ}} & = \underbrace{\frac{\sum_{(R, P)\in \text{TP}} \text{\textit{f}}(R, P)}{\vert \text{TP} \vert}}_{\text{\textit{Segmentation quality (SQ)}}} \cdot \underbrace{\frac{\vert \text{TP} \vert}{\vert \text{TP} \vert + 0.5 \cdot \vert \text{FP} \vert + 0.5 \cdot \vert \text{FN} \vert}}_{\text{\textit{Recognition quality (RQ)}}}
    \end{aligned}
    \label{equation:pq_separate}
\end{equation}

where $R$ refers to the reference instances and $P$ to the predicted instances, and $f$ is the similarity metric.

In the original paper \citet{kirillov2019panoptic} propose to use \eac{IOU} as similarity metric, while \textit{panoptica} also implements \eac{PQ} based on \eac{DSC} and \eac{ASSD} in addition to \eac{IOU}.
Note that the code can easily be modified to integrate further desired metrics, \cf \Cref{sec:methods}.

\noindent\textbf{Edge case handling:}
For some edge cases, segmentation metrics are not defined.
For example, if both reference and prediction are empty \eac{DSC} and \eac{ASSD} become \textit{NaN}, however, some might argue that \eac{DSC} should be defined as $1.0$ for this case.
A similar problem occurs for distance-based metrics such as \eac{ASSD}, which become \textit{Inf} if either prediction or reference is empty.
\textit{NaN} and \textit{Inf} make it challenging to report aggregated metrics such as the \textit{mean}.
Therefore, practitioners developed task-specific mitigation strategies such as assigning a theoretically possible maximum distance for distance-based metrics \citep{bakas2019identifying}.
In the absence of a general mitigation strategy producing satisfying results across all use cases, \textit{panoptica} equips users with control over edge case handling.

\clearpage
\section{Evaluation experiments}
\label{sec:experiments}
We conduct three biomedical evaluation experiments to showcase \textit{panoptica} for various use cases.
Across all experiments, we report the \eacf{gvDSC} and the instance-wise \eacf{RQ}, \eacf{SQ}, \eacf{PQ} and \eacf{SQASSD}.
Note that \eac{SQASSD} becomes \textit{NaN} for cases with empty prediction and reference, and \textit{Inf} for cases with either empty prediction or empty reference or only unmatched instances, \cf \Cref{sec:evaluator}.
Therefore, our mean and standard deviation cases are only based on cases that return a number.


\newcolumntype{Y}{>{\centering\arraybackslash}X}

\subsection{VerSe experiment}
\label{sec:verse}
The \eac{VerSe} challenge is about localization, labeling, and segmentation of vertebrae in CT scans of arbitrary fields of view \citep{sekuboyina2021verse}.
For the segmentation part, the references were \textit{instance maps} with labels that correspond to their anatomical label (i.e., 20 for L1).
A simple volumetric \eac{DSC} calculated on a binary version of this mask shows an overall semantic segmentation performance.
However, instance-wise metrics are needed to measure how well an algorithm can actually distinguish the different vertebrae.
Additionally, as in some cases, the actually different instance labels may be deemed relevant.
If a model segments everything correctly but has all labels shifted by one, the resulting overall semantic similarity metric would be terrible, as this is not taken into account.
\textit{panoptica} allows the calculation of both metrics dynamically by stating whether the labels of the input maps are assumed to be already matched or not.

\begin{figure}[H]
  \centering
  \includegraphics[width=0.94\columnwidth]{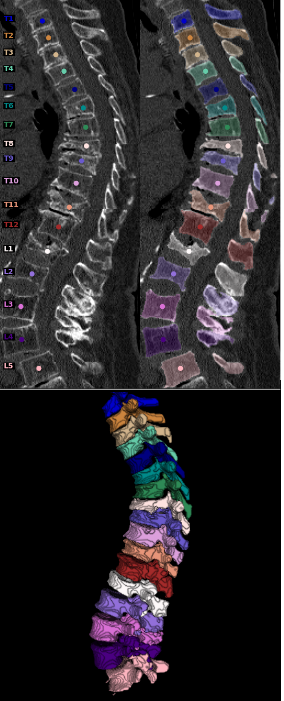}
  \caption{
    The \eac{VerSe} segmentation challenge task.
    The reference segmentations for the challenge were generated through a hybrid approach, first being automatically segmented and then manually corrected.
    This figure shows an exemplary \eac{CT} scan (top-left) of the test set, its reference segmentation mapped on the image (top-right), and a 3D rendering of the reference segmentation (bottom).
  }
  \label{fig:verse_task}
\end{figure}

\noindent\textbf{Segmentation Task:}
The task in the challenge is that of multi-class instance segmentation, though mainly tackled as a semantic segmentation task by the participants, \cf \Cref{fig:verse_task}.

\noindent\textbf{Data:}
The dataset features \eac{CT} scans of various scanners, resolutions, and fields of view.
We evaluate the segmentation algorithms on the hidden test set of the challenge consisting of $103$ subjects.
The number of instances can vary between $5$ and $25$ (mean $13.08\pm5.416$) per scan.

\noindent\textbf{Procedure:}
We use the top three submissions from the \eac{VerSe} challenge and evaluate their predictions with the corresponding reference maps.
We consider the input to \textit{panoptica} to be unmatched instance maps, as we are not interested in evaluating the labeling capabilities but the segmentation metrics irrespective of their actual predicted vertebra labels.
\Cref{tab:verse_table} illustrates the results of our analysis.

\begin{table*}[htbp]
    \centering
    \begin{tabularx}{\textwidth}{XYcccc}
        \toprule
                                         & Global Metrics         & \multicolumn{4}{c}{Instance-wise Metrics}                                                                               \\
        \cmidrule(lr){2-2}
        \cmidrule(lr){3-6}
        %
        Submission                       & \eac{gvDSC} $\uparrow$ & \eac{RQ} $\uparrow$                       & \eac{SQ} $\uparrow$    & \eac{PQ} $\uparrow$    & \eac{SQASSD} $\downarrow$ \\
        \midrule
        Payer et al. \citep{visapp20}    & 0.932 ± 0.041          & 0.97 ± 0.046                              & \textbf{0.912} ± 0.028 & \textbf{0.886} ± 0.048 & \textbf{0.37} ± 0.055     \\
        deepreasoningai \citep{chen20}   & \textbf{0.937} ± 0.036 & \textbf{0.982} ± 0.038                    & 0.898 ± 0.046          & 0.882 ± 0.063          & 0.414 ± 0.098             \\
        jdlu \citep{sekuboyina2021verse} & 0.901 ± 0.037          & 0.947 ± 0.058                             & 0.865 ± 0.039          & 0.818 ± 0.059          & 0.601 ± 0.41              \\
        \bottomrule
    \end{tabularx}
    \caption{Performance metrics for the top-three submissions of the \eac{VerSe} challenge using an \eac{IOU} threshold of $0.5$ for both matching and evaluation.
    Besides \eac{gvDSC}, we report instance-wise metrics, \cf \Cref{sec:experiments}.
    Mean ± standard deviation values are reported.
    Metrics with $\uparrow$ are better, the higher the value, while $\downarrow$ is the opposite.
    The best values for each metric are highlighted in bold.
    Each algorithm performs quite well. However, although the submission from \citet{chen20} has the overall best \eac{gvDSC}, it lacks behind \citet{visapp20} in \eac{SQ}, \eac{PQ}, and \eac{SQASSD}. This suggests the model from \citet{chen20} is better performing in detecting the correct instances, while the model from \citet{visapp20} is better at segmenting the instances it correctly detected.
    }
    \label{tab:verse_table}
\end{table*}

\subsection{ISLES experiment}
\label{sec:isles}
The \eac{ISLES} experiment serves as a practical example for the instance-wise evaluation of semantic segmentation maps.
The segmentation of ischemia, addressed as a \eac{SemS} task in \eac{ML}, is the focus of the ISLES'22 challenge.
Despite this, the task and associated data encompass a wide variability of instances (lesions) requiring accurate delineation.
Consequently, instance-wise analysis is valuable for tackling the problem.
Specifically, the task involves scans featuring multiple non-overlapping embolic brain lesions that exhibit significant variations in terms of size, quantity, and anatomical location.

\begin{figure}[htbp]
  \centering
  \includegraphics[width=1.0\columnwidth]{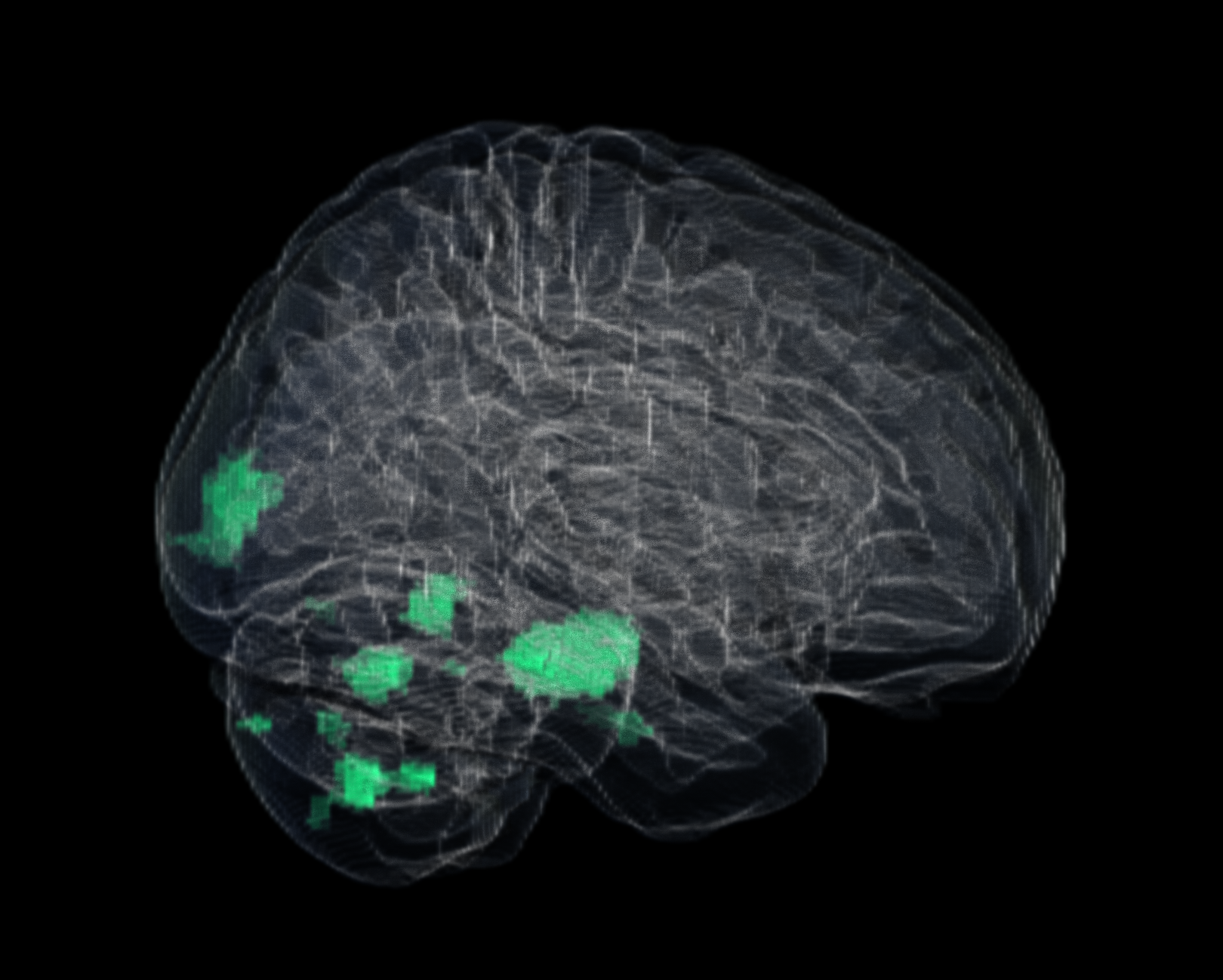}
  \caption{
    A 3D-rendered human brain from a sample scan in the \eac{ISLES} dataset, with binary stroke lesions overlaid in green.
    Participants in the \eac{ISLES} challenge are tasked with developing algorithms for segmenting ischemic brain lesions based on multi-modal \eac{MRI} data.
    As depicted in the image, these lesions can be multiple, non-overlapping, and distributed across various locations in the brain.
    }
  \label{fig:isles_task}
\end{figure}

\noindent\textbf{Segmentation Task:}
The \eac{ISLES} challenge features a binary semantic segmentation task illustrated in \Cref{fig:isles_task}.
The goal is to segment ischemic stroke brain lesions from multi-modal MRI data.
Unlike this work, the original challenge evaluates the submissions' performance using \eac{gvDSC}, \textit{F1-score} as \textit{instance-wise} (here equals lesion-wise) detection, and further clinical metrics such as the differences in absolute lesion volume and absolute lesion count.

\noindent\textbf{Data:}
We analyze the test set of the \href{https://isles22.grand-challenge.org/}{ISLES'22 MICCAI challenge} \citep{hernandez2022isles}.
The dataset consists of $150$ multi-center, multi-scanner MR scans.
Each exam features three \eac{MRI} sequences, namely \eacf{DWI}, \eacf{ADC}, and \eacf{FLAIR}, accompanied by a label mask encoding the stroke lesions.
The data includes a wide variability of image acquisition and stroke patterns.
The reference segmentations are obtained through an iterative hybrid algorithm-expert approach and are double-checked by experienced neuroradiologists.
The number of non-overlapping instances (i.e., lesions) varies from none (i.e., healthy scans) to a hundred-twenty-six \citep{hernandez2022isles}.

\noindent\textbf{Procedure:}
In this work, we conduct a complementary analysis to the one conducted in the ISLES'22 challenge by assessing the top-three ranked methods with instance-wise metrics obtained through \textit{panoptica}.
We evaluate the submissions from team \textit{seals} (rank $\#$1) \citep{dockerSeals}, \textit{nvauto} (rank $\#$2) and \href{https://isles22.grand-challenge.org/}{swan} (rank $\#$3).
Team \textit{seals} employs an ensemble of \textit{nnU-Nets} \citep{isensee2021nnu}; team \textit{nvauto} applies the \textit{Auto3dSeg} pipeline \citep{siddique2022automated} from \eac{MONAI} \citep{cardoso2022monai}, while the team \textit{swan} utilizes a non-negative matrix factorization neural network \citep{ashtari2023factorizer}.
Performance metrics obtained for each of these algorithms are summarized in Table \Cref{tab:isles_table}.

\begin{table*}[htbp]
    \centering
    \begin{tabularx}{\textwidth}{XYcccc}
        \toprule
         & Global Metrics & \multicolumn{4}{c}{Instance-wise Metrics}\\
        \cmidrule(lr){2-2}
        \cmidrule(lr){3-6}
        %
        Submission                    & \eac{gvDSC} $\uparrow$                   & \eac{RQ} $\uparrow$                  & \eac{SQ} $\uparrow$                  & \eac{PQ} $\uparrow$                   & \eac{SQASSD} $\downarrow$                \\
        \midrule
        seals \citep{dockerSeals}  & 0.781 ± 0.181 & \textbf{0.52} ± 0.285          & \textbf{0.669} ± 0.217 & \textbf{0.380} ± 0.215 & 0.312 ± 0.175 \\
         nvauto \citep{siddique2022automated} & \textbf{0.783} ± 0.177          & 0.485 ± 0.278 & \textbf{0.669} ± 0.217          & 0.357 ± 0.211 & \textbf{0.307} ± 0.166          \\
        swan \citep{ashtari2023factorizer}                          & 0.763 ± 0.194          & 0.462 ± 0.285          & 0.634 ± 0.253          & 0.334 ± 0.213          & 0.353 ± 0.292          \\
        \bottomrule
    \end{tabularx}
    \caption{
    Performance metrics for the top-three ranked \eac{ISLES}'22 challenge algorithms computed with an \textit{IoU} threshold of $0.5$  for both matching and evaluation.
    Besides \eac{gvDSC}, we report instance-wise metrics, \cf \Cref{sec:experiments}.
    Mean ± standard deviation values are reported.
    Metrics with $\uparrow$ are better, the higher the value, while $\downarrow$ is the opposite.
    The best values for each metric are highlighted in bold.
    Statistics for \eac{SQASSD} were performed disregarding infinite values, which involved 13 samples from the teams \textit{seals} and \textit{nvauto} and 19 samples from the team \textit{swan}.
    Results show that team \textit{nvauto} provided an overall better semantic segmentation performance in terms of $gvDSC$, while the team \textit{seals} achieved better segmentation performance at the instance level.}
    \label{tab:isles_table}
\end{table*}


\newpage
\subsection{BraTS Mets experiment}
\label{sec:brats}
The \textit{BraTS Mets} experiment represents a use case for instance-wise evaluation of semantic segmentation maps.
Although metastasis segmentation is an instance segmentation task from a clinical perspective, similar to stroke segmentation \cf \Cref{sec:isles}, it is typically operationalized as a \eac{SemS} task in \eac{ML}, \cf \Cref{sec:introduction}.

\begin{figure}[H]
  \centering
  \includegraphics[width=0.9\columnwidth]{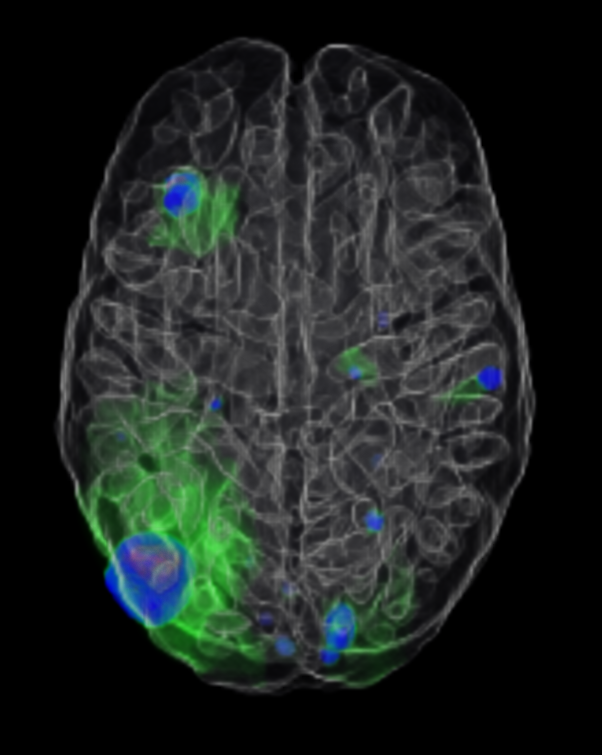}
  \caption{
    The \eac{BraTS-M} hierarchical semantic segmentation task - axial view of 3D rendered brain with overlayed colored segmentation.
    The \eac{BraTS-M} \citep{moawad2023brain} segmentation task is inspired by the adult glioma \href{https://www.synapse.org/\#!Synapse:syn51156910/wiki/622351}{\eac{BraTS-G}} \citep{bakas2019identifying}.
    The glioma lesions are segmented with the classes \eac{ET} (blue), \eac{NETC} (red), and \eac{SNFH} (green).
    However, the algorithms are trained and evaluated on hierarchical class channels building upon each other, namely, first, the \textit{enhancing tumor} channel, which consists only of such.
    Furthermore, the \textit{tumor core} channel combines \eac{ET} and \eac{NETC}, and the \textit{whole tumor} channel combines all three classes.
    The instance-wise evaluation presented in \Cref{tab:brats_table} considers only the \textit{tumor core} channel.
  }
  \label{fig:brats_task}
\end{figure}

\noindent\textbf{Segmentation Task:}
The original segmentation task of the \href{https://www.synapse.org/#!Synapse:syn51156910/wiki/622553}{\eac{BraTS-M} \citep{moawad2023brain}} is a multi-class hierarchical semantic segmentation task, \cf \Cref{fig:brats_task}.

\noindent\textbf{Data:}
We evaluate on the publicly available training set of the \eac{BraTS-M} \citep{moawad2023brain}.
The dataset consists of $726$ examinations, each containing four \eac{MRI} sequences, namely a \eac{t1w}, a \eac{t1c}, a \eac{t2w}, and a \eacf{t2f} sequence combined with a multi-class semantic segmentation label, \cf \Cref{fig:brats_task}.
Metastases counts in the dataset hugely vary between $1$ and $393$ (mean $8.68\pm18.64$) per exam.

\noindent\textbf{Procedure:}
To explore how the glioma segmentation algorithms are able to generalize to cancer metastasis segmentation, we conduct the following \textit{off-label use experiment}.
We segment the whole dataset with five publicly available glioma \eac{SemS} algorithms distributed through \href{https://github.com/neuronflow/BraTS-Toolkit}{BraTS Toolkit} \citep{kofler2020brats}, namely \textit{hnfnetv1-20} \citep{jia2021hnfnetv1}, \textit{isen-20} \citep{isensee2021isen}, \textit{sanet0-20} \citep{yuan2021sanet}, \textit{scan\_lite-20} \citep{mckinley2021scan_lite}, and \textit{yixinmpl-20} \citep{wang2021yixinmpl}.
Additionally, we derive a sixth segmentation map named \textit{simple} by fusing the five previously obtained algorithms using the implementation of the \textit{SIMPLE} \citep{langerak2010simple} fusion algorithm in BraTS Toolkit \citep{kofler2020brats}.
We then compute the \eac{gvDSC} and the instance-wise metrics using \textit{panoptica}, \cf \Cref{sec:experiments}.
The results of this analysis are illustrated in \Cref{tab:brats_table}.

\begin{table*}[htbp!]
    \centering
    \begin{tabularx}{\textwidth}{XYcccc}
        \toprule
                                                    & Global Metrics         & \multicolumn{4}{c}{Instance-wise Metrics}                                                                           \\
        \cmidrule(lr){2-2}
        \cmidrule(lr){3-6}
        %
        Submission                                  & \eac{gvDSC} $\uparrow$ & \eac{RQ} $\uparrow$                       & \eac{SQ} $\uparrow$  & \eac{PQ} $\uparrow$  & \eac{SQASSD} $\downarrow$ \\
        \midrule
        hnfnetv1-20 \citep{jia2021hnfnetv1}         & 0.52 ± 0.39            & 0.33 ± 0.35                               & 0.50 ± 0.39          & 0.27 ± 0.28          & 0.65 ± 0.37               \\
        isen-20 \citep{isensee2021isen}             & \textbf{0.54} ± 0.39   & \textbf{0.37} ± 0.36                      & 0.52 ± 0.38          & \textbf{0.29} ± 0.30 & \textbf{0.64} ± 0.32      \\
        sanet0-20 \citep{yuan2021sanet}             & 0.53 ± 0.38            & 0.35 ± 0.35                               & \textbf{0.53} ± 0.38 & 0.28 ± 0.29          & 0.65 ± 0.35               \\
        scan\_lite-20 \citep{mckinley2021scan_lite} & 0.49 ± 0.37            & 0.32 ± 0.35                               & 0.46 ± 0.38          & 0.25 ± 0.28          & 0.70 ± 0.38               \\
        yixinmpl-20 \citep{wang2021yixinmpl}        & 0.52 ± 0.38            & 0.34 ± 0.35                               & 0.50 ± 0.37          & 0.27 ± 0.28          & 0.70 ± 0.34               \\
        simple \citep{langerak2010simple}           & 0.52 ± 0.39            & 0.35 ± 0.36                               & 0.51 ± 0.38          & 0.28 ± 0.29          & 0.68 ± 0.36               \\
        \bottomrule
    \end{tabularx}
    \caption{
        Table summarizing the \textit{BraTS Mets experiment} for an \eac{IOU} threshold of $0.5$ for both matching and evaluation.
        Besides \eac{gvDSC}, we report instance-wise metrics, \cf \Cref{sec:experiments}.
        Mean ± standard deviation values are reported.
        Metrics with $\uparrow$ are better, the higher the value, while $\downarrow$ is the opposite.
        The best values for each metric are highlighted in bold.
        We observe that all algorithms miss many metastases as reflected by the low \eac{RQ}.
        Further, the \eac{SQ} and \eac{SQASSD} for the correctly detected metastases are only moderate, resulting in an overall low \eac{PQ}.
        Since \eac{SQASSD} is only a number for patients with matched instances (\cf \Cref{sec:evaluator}), samples have been excluded from the calculation of this metric (from $238$ for \textit{sanet0-20} up to $286$ for \textit{scan\_lite-20}).
        The low \eac{PQ} and \eac{gvDSC} indicate that the glioma algorithms are not able to generalize well to cancer metastasis segmentation.
        This is even worse than the moderate \eac{gvDSC} suggests, as indicated by the discrepancy between \eac{PQ} and \eac{gvDSC}. 
    }
    \label{tab:brats_table}
\end{table*}




\section{Discussion}
\label{sec:discussion}
\textit{panoptica} is an open-source, performance-optimized package written in Python.
It allows the computation of instance-wise segmentation quality metrics, for instance- and semantic segmentation maps.
Especially in the biomedical domain, for which we demonstrate three example use cases, instance-wise evaluation can be crucial to achieving a better representation of the underlying clinical task.
Beyond our examples, there are countless segmentation problems, such as cell counting, that can profit from an instance-wise evaluation.
As this also applies to the segmentation of natural images, \textit{panoptica} is not limited to applications in the biomedical domain.

\noindent\textbf{Limitations and future work:}
Transforming semantic segmentation problems into instance segmentation outputs via \eac{CCA} comes with the drawback of typically discarding predicted class scores by relying on integer values as label maps.
However, predicted class scores can be important and relevant for the validation.
Luckily, instance-wise class scores can be retained from the \eac{SemS} as shown in \cite{reinke2021common}.

The metrics provided by \textit{panoptica} can be easily extended to further metrics such as more selection for overlap- and boundary-based metrics, for example, by including \eac{clDice}, \eac{NSD}, or \eac{HD}.
In addition, the package could be extended to include multi-threshold metrics \cite{maierhein2023metrics} such as \eac{AP} and \citep{chen2023sortedap} to assess detection quality in more detail.
With a more comprehensive list of metrics, the package will be able to offer a  more extensive performance assessment from different viewpoints and allow users to select the most suitable choices for their application.

\newpage
The current package implements several overlap-based strategies.
Another worthy addition would be to implement more matching strategies to cover a bigger variety of use cases.
For example, if predicted class scores are available for each instance, the \textit{Greedy by Score} matching \cite{everingham2015pascal} could be applied.
This strategy orders predicted instances by their predicted class scores and identifies the matching \textit{reference instance} by choosing the \textit{reference instance} with the highest \eacf{MS}.
Next, the \textit{assigned reference instance} will be removed, such that this instance matcher remains a one-to-one matcher.
If class scores are not available, the strategy can be simplified to the \textit{Greedy by Localization Criterion} matching \cite{maierhein2023metrics}, which orders predicted instances by the \eac{MS}.
Furthermore, more advanced instance matchers like the \textit{Optimal Hungarian Matching} \cite{kuhn1955hungarian} can be applied, which minimizes a cost function dependent on the \eacf{MM}.
Another interesting idea would be to implement matching based on topological features.

Similarly, we aim to extend the type of \eac{MM} to include all localization criteria mentioned in the Metrics Reloaded framework \cite{maierhein2023metrics}, as it has been shown \cite{maierhein2023metrics} that \ac{IOU} is not always the best option for a localization criterion.
For a comprehensive analysis of performance variability, we aim to expand the notion of mean and standard deviation to confidence intervals, for example, constructed via bootstrapping.






\clearpage

\clearpage
{
    \small
    \bibliographystyle{ieeenat_fullname}
    \bibliography{references}
}

\clearpage
\clearpage
\onecolumn
\setcounter{section}{41}
\section{Supplemental Material}
\label{sec:supplement}

\subsection{\eac{CCA} Benchmark: \textit{cc3d} vs. \textit{scipy}}
\label{sec:benchmark}

\Cref{tab:scipy_cc3d_benchmark} summarizes the comparison of \eac{CCA} backends. The code for the comparison is \href{https://github.com/BrainLesion/panoptica/blob/main/benchmark/benchmark.py}{publicly available}.
\begin{table}[ht]
  \centering
  \begin{tabularx}{0.56\textwidth}{Xrrr}
    \toprule
    Dimensionality & Volume Size & Scipy Time [s]  & CC3D Time [s]     \\
    \midrule
    2D             & \(500^2\)   & \textbf{0.0350} & 0.0552            \\
                   & \(1000^2\)  & \textbf{0.1344} & 0.2268            \\
                   & \(2000^2\)  & \textbf{0.5121} & 0.8542            \\
                   & \(4000^2\)  & \textbf{2.1542} & 3.4198            \\
                   & \(8000^2\)  & \textbf{9.5657} & 16.4068           \\
    \midrule
    3D             & \(50^3\)    & 0.0204          & \textbf{0.0144}   \\
                   & \(100^3\)   & 0.1564          & \textbf{0.1160}   \\
                   & \(200^3\)   & 1.2277          & \textbf{0.9043}   \\
                   & \(500^3\)   & 20.7657         & \textbf{15.3032}  \\
                   & \(1000^3\)  & 195.1927        & \textbf{145.1652} \\
    \bottomrule
  \end{tabularx}

  \caption{
    Benchmarking Scipy's skimage vs. cc3d for connected component analysis for random binary 2D and 3D data.
    The benchmark suggests that Skimage is more efficient for 2D data, while cc3d performs better for 3D data.
    The tests were conducted on an Apple M1 Pro Processor.
    \textit{panoptica} comes with \href{https://github.com/BrainLesion/panoptica/blob/f327a0f3ecde57eddeb7643e0a67d726b96b1c18/benchmark/benchmark.py}{code} to benchmark which backend performs better on the given hardware.
  }
  \label{tab:scipy_cc3d_benchmark}
\end{table}

\subsection{Computation times}
\label{sec:times}
\begin{figure}[H]
  \centering
  \includegraphics[width=1.0\columnwidth]{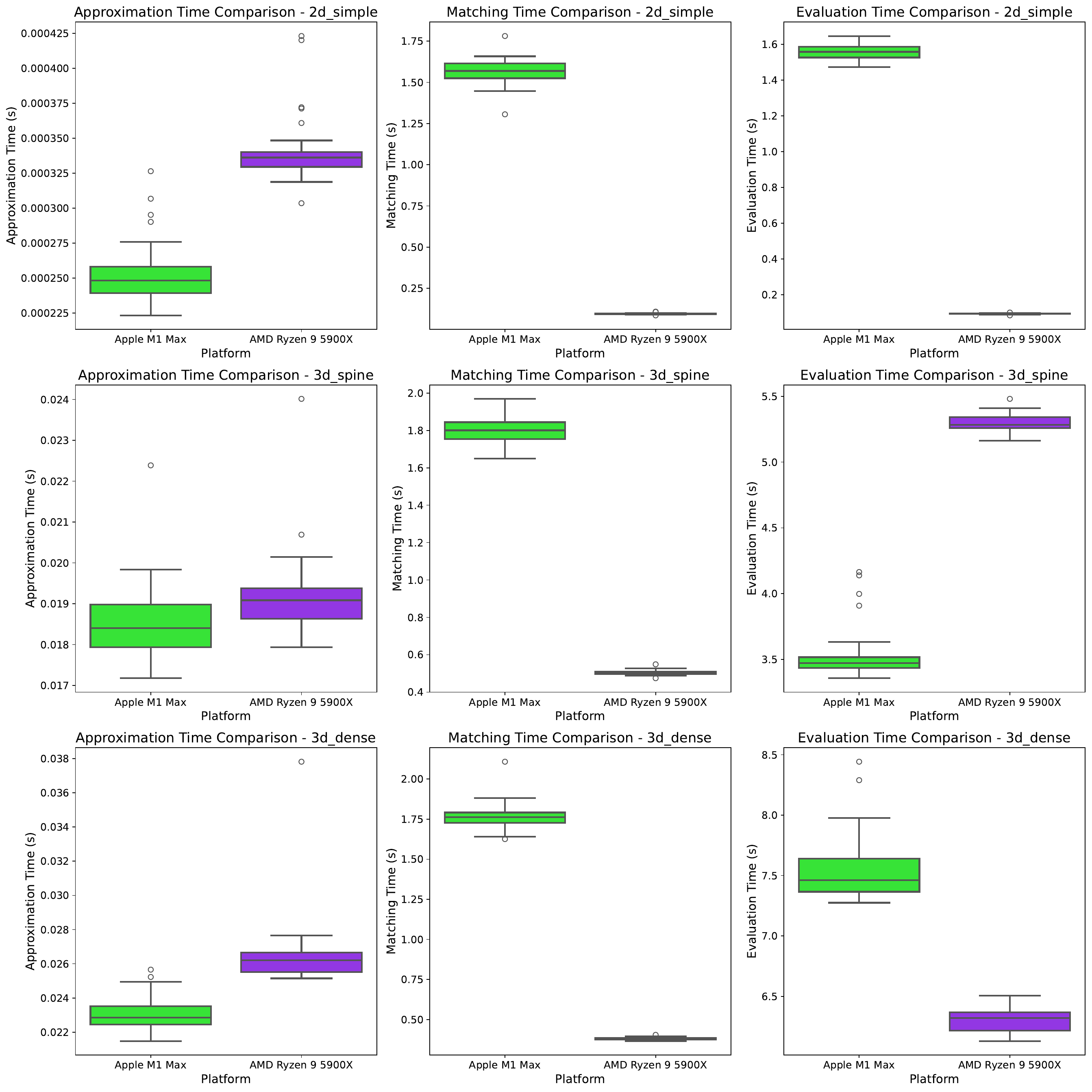}
  \caption{
    Performance test of the \textit{panoptica} package over two different machines (green and purple) and three unique input samples (rows).
    Each test shows the mean and standard deviations across 42 iterations for each of the three modules (columns).
    All measurements are given in seconds.
    While instance approximation is consistently fast, instance matching and evaluation can take a couple of seconds, depending on the input sizes and parallel processing power of the utilized CPU.
    As \textit{panoptica} allows the user to specify which metrics should be calculated, these times can vary.
  }
  \label{fig:boxplot_times}
\end{figure}

We benchmark the three modules from \textit{panoptica} on exemplary samples.
Across all these performance tests, we use instance approximation based on \eac{CCA}, instance matching with the naive one-to-one setting (\cf \Cref{sec:matching}) using \eac{IOU} as \eac{MM}, and evaluate the instance-wise metrics \eac{RQ}, \eac{SQ}, \eac{PQ} and \eac{ASSD}.
For all experiments, we use a \eac{IOU} threshold of $0.5$ both for matching and evaluation.
The following test setups are used:
\begin{itemize}
    \item Work station: Ubuntu 22.04, AMD Ryzen 9 5900x, 128GB RAM
    \item Apple Macbook Pro with M1 Max 10-Core processor, 32GB RAM
\end{itemize}
The code to run the benchmark is \href{https://github.com/BrainLesion/panoptica/blob/main/benchmark/modules_speedtest.py}{publicly available}.
The exemplary samples include:
\begin{itemize}
    \item 2d\_simple: A simulated (50,50) 2D input consisting of two instances both in prediction and reference.
    \item 3d\_spine: A real MRI spine segmentation sample of shape (170, 512, 17) consisting of 89 prediction instances and 87 reference instances.
          73 instances are successfully matched.
    \item 3d\_dense: A (100, 100, 100) shaped image densely filled with 40 simulated instances both in prediction and reference.
\end{itemize}
\Cref{fig:boxplot_times} shows the result of our tests.

\clearpage
\subsection{Segmentation Performance over varying \eac{IOU}-thresholds}
\label{sec:curves}

\begin{figure}[H]
  \centering
  \includegraphics[width=1.0\columnwidth]{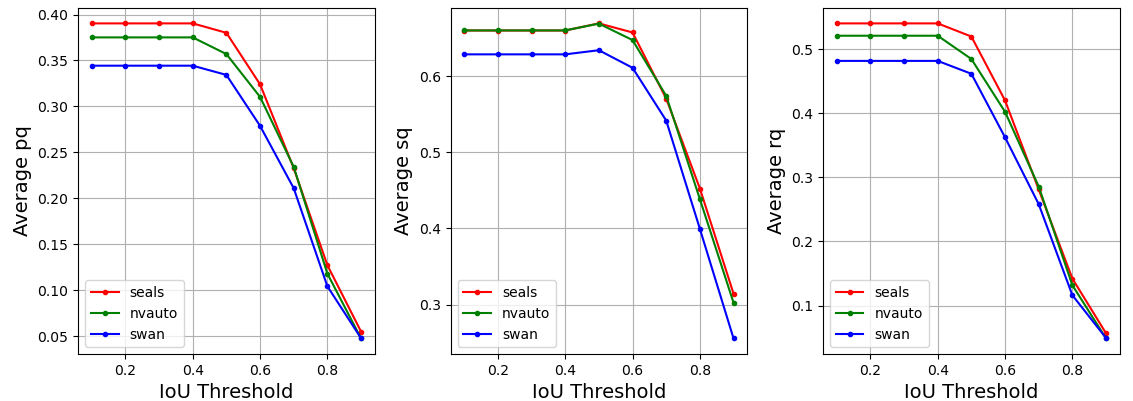}
  \caption{
  Average \eac{PQ} (left), -\eac{SQ} (middle), -\eac{RQ} (right) for varying \eac{IOU}-thresholds.
  Here, the same threshold is applied for both \textit{instance matching} and \textit{instance evaluation}.
  Each curve represents one of the leading teams of the \eac{ISLES} challenge, \cf \Cref{sec:isles}.
  Notably, depending on the IOU threshold and employed metric,  it varies which is the optimal algorithm for stroke segmentation.
  One could consider utilizing the \eacf{AUC} for each algorithm across the entire range of thresholds to provide a holistic performance summary robust to variations in threshold selection.
  }
  \label{fig:iou_curves}
\end{figure}

With \textit{panoptica} performance measures can easily be computed for various \eac{IOU}-thresholds, \cf \Cref{fig:iou_curves}.
Note that different thresholds could be selected for \textit{instance matching} and \textit{instance evaluation}.

\section*{Acknowledgement}
Bjoern Menze, Benedikt Wiestler, and Florian Kofler are supported through the SFB 824,
subproject B12. Supported by Deutsche Forschungsgemeinschaft (DFG) through TUM International Graduate School of Science and Engineering (IGSSE), GSC 81.
Supported by Anna Valentina Lioba Eleonora Claire Javid Mamasani.
Bjoern Menze acknowledges support by the Helmut Horten Foundation.
Ivan Ezhov is supported by the Translational Brain Imaging Training Network (TRABIT) under the European Union’s ‘Horizon 2020’ research \& innovation program (Grant agreement ID: 765148).
Fabian Isensee is supported by Helmholtz Imaging (HI), a platform of the Helmholtz Incubator on Information and Data Science.
Jan Kirschke has received Grants from the ERC, DFG, BMBF and is a Co-Founder of Bonescreen GmbH.
Benedikt Wiestler is supported by the NIH (R01CA269948).


\end{document}